\documentclass[conference]{IEEEtran}
\IEEEoverridecommandlockouts
\usepackage{cite,url,hyperref}
\usepackage{amsmath,amssymb,amsfonts}
\usepackage{algorithmic}
\usepackage{graphicx}
\usepackage{textcomp}
\usepackage{xcolor}
\def\BibTeX{{\rm B\kern-.05em{\sc i\kern-.025em b}\kern-.08em
    T\kern-.1667em\lower.7ex\hbox{E}\kern-.125emX}}
\begin{document}
\title{FOXNET: A MULTI-FACE ALIGNMENT METHOD}
\author{\IEEEauthorblockN{
Yuxiang Wu$^{1}$, Zehua Cheng$^2$, Bin Huang$^2$, Yiming Chen$^{1,\star}$\thanks{$^\star$Corresponding authors.}, Xinghui Zhu$^{1}$,Weiyang Wang$^{2, \star}$\\
\textit{$^1$AISA Research, Hunan Agriculture University}, Changsha, China\\
\{wyxiang,chenym,zhxh\}@hunau.edu.cn\\
\textit{$^2$SnowCloud.ai}, Beijing, China\\
\{limber,bin.huang,weiyang.wang\}@snowcloud.ai\\
}}
\maketitle
\begin{abstract}
    Multi-face alignment aims to identify geometry structures of multiple faces in an image, and its performance is essential for the many practical tasks, such as face recognition, face tracking, and face animation. 
    In this work, we present a fast bottom-up multi-face alignment approach, which can simultaneously localize multi-person facial landmarks with high precision.
    In more detail, our bottom-up architecture maps the landmarks to the high-dimensional space with which landmarks of all faces are represented. By clustering the features belonging to the same face, our approach can align the multi-person facial landmarks synchronously.
    Extensive experiments show that our method can achieve high performance in the multi-face landmark alignment task while our model is extremely fast. 
    Moreover, we propose a new multi-face dataset to compare the speed and precision of bottom-up face alignment method with top-down methods.
    Our dataset is publicly available at~\footnote{\url{https://github.com/AISAResearch/FoxNet}}. 

    % Face alignment aims to identify geometry structures of human face, and the alignment performance is important for the many practical tasks, such as face recognition, face tracking and face animation. Multi-face alignment is one of the challenging tasks in face alignment, and has been extensively used in various applications. In this work, we present a fast facial landmark detection approach, which can simultaneously localize multi-person facial landmarks with high precision. In more detail, unlike previous top-down approach, our method uses a bottom-up architecture to map the landmarks to the high-dimensional space.
    % Then, the discriminative high-dimensional features are aggregated to represent the landmarks. By clustering the aggregated features belonging to the same face, our approach can align the multi-person facial landmarks synchronously.
    % Extensive experiments are conducted in this paper, and the experimental results demonstrate that our method can achieve the high performance in the multi-face landmark alignment task while our model is extremely fast. 
    % Moreover, we propose a new multi-face dataset to compare the speed and precision of bottom-up face alignment method.
    % Our dataset is publicly available at~\footnote{\url{https://github.com/AISAResearch/FoxNet}}. 
\end{abstract}


\section{Introduction}
Deep learning has made great progress in recent days; one of the most compelling achievements is the application of computer vision.
Multi-face alignment, also known as multiple facial landmarks localization or detection, aims to identify the locations of the key points of multiple faces on images or videos.

% \begin{enumerate}
%     \item Directly finding the landmarks using deep convolutional neural networks structures: This approach has a strong correlation with the face dataset where the label rules and the number of landmarks are different, and it is required to retrain the entire networks while using another dataset~\cite{shao2018deep}.
%     \item Indirectly utilizing the landmark information: A curve connected by landmarks, inaccurate supervision, is introduced to parse the corresponding landmarks~\cite{wu2018look}.
%      The generated boundary is independent of the number of the landmarks by the dataset. Therefore, any number of a landmark could be generated using the supervised information provided by the face boundaries.
% \end{enumerate}
\begin{figure*}
    \centering
    \includegraphics[scale=0.5]{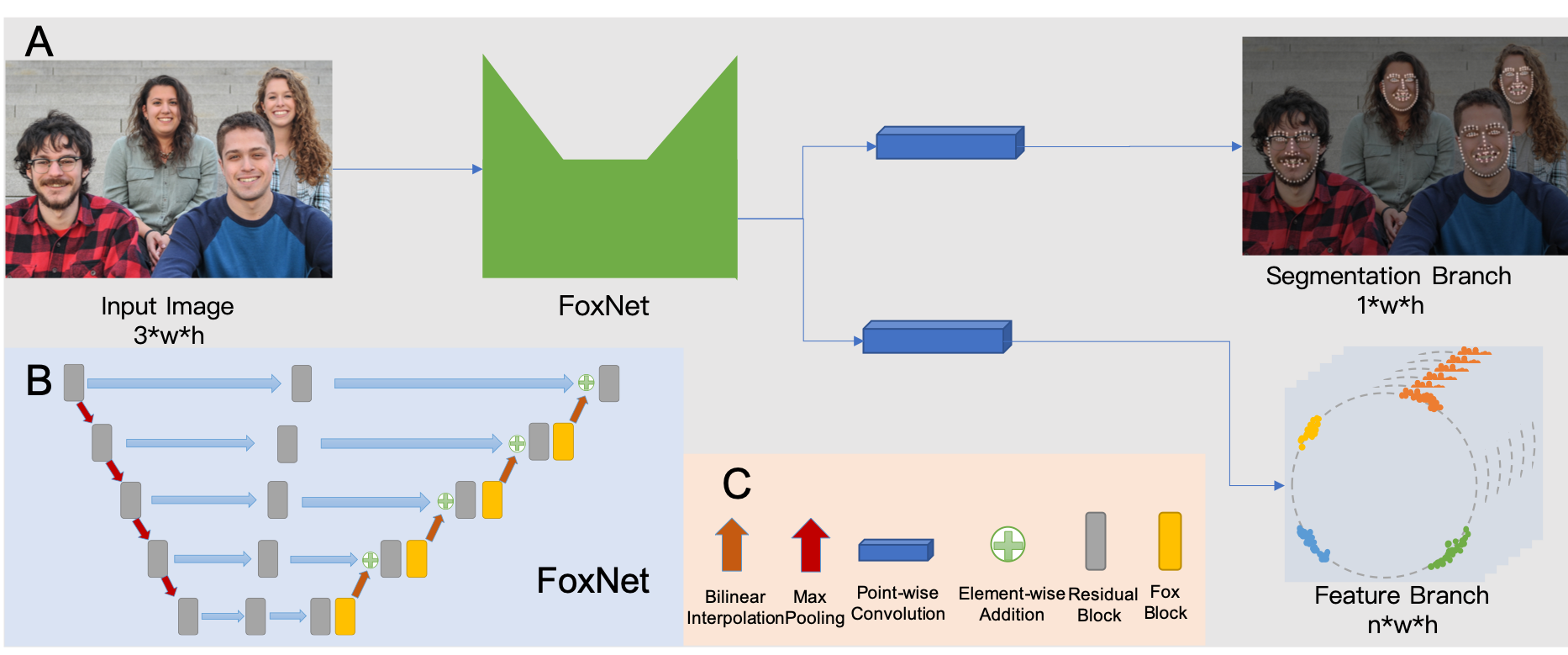}
    \caption{Overview of our proposed model training structure (A) and our proposed FoxNet (B). 
    (A) For a given image, we first use a ResNet50 and several FoxBlock to extract to the feature map, and utilize
    two points-wise convolution to get the landmark candidates and pixel embeddings. 
    (B) We use our Fox Block before skip-connection.
    \label{fig:model_structure}}
\end{figure*}
Multi-face alignment task can be grouped into bottom-up and top-down approaches. 
For a long time in academia and industry, people have employ a top-down method, face detection first and then send to single face alignment network.
Conventional single face alignment methods ~\cite{farfade2015multi,ranjan2015deep,yang2015facial,luo2012hierarchical,sun2013deep} can be divided into directly or indirectly generating landmarks~\cite{wu2018look}.
% There are more top-down based methods to utilize global context and strong structural information in this task since it required more details to maintain accuracy.
%are more robust to occlusion and complex poses than top-down. What is more, their 
The time complexity of NMS and Soft NMS~\cite{bodla2017soft} is $O(n^{2})$, which is the most critical deficiency of the top-down method. After that, the results of NMS or Soft NMS ~\cite{bodla2017soft}are sent to the single face detection network~\cite{hu2017finding,tang2018pyramidbox,jiang2017face,wang2017detecting,zhang2017s}. 
For this process, the time complexity is $O(n)$. What is worse, traditionally, the single face detection networks have a very deep convolution structure. Repeated use of convolutional networks to infer images can greatly slow down the entire structure.
% From the perspective of directly or indirectly using landmark information, the top-down structure has two different approaches to achieve single face alignment tasks.
% However, to achieve a multi-face alignment task (multiple faces in one picture), the top-down method has sacrificed the speed to get better accuracy. 
% The top-down approach detects multiple faces first, then send each detected faces to the single face alignment networks to generate the individual face sequentially.
% This approach has divided the task into detecting the single face problem.
% Because the top-down structure passes each detected face to the face landmark localization networks, 
Once the number of faces increases, the speed will be greatly sacrificed. 

So, it is important to develop a bottom-up structure for multi-face alignment task.
% The runtime of the face detection networks is proportional to the shape of the image, NMS and the number of faces, single face key points are not limited by the shape of the image and the number of faces, but every face needs to pass it, so In general, the step is proportional to the number of faces and the shape of the image. 
Some bottom-up human pose estimation algorithms~\cite{wei2016convolutional,cao2018openpose,cao2017realtime} use Part Affinity Fields and a greedy parse to resolve individual. Inspired by that, our multi-face bottom-up method can be divided into two steps~\cite{cao2017realtime}: first finding out all possible face landmarks, and then parsing the discrete key points into individuals. 
Since this method is based on the entire image, it needs to overlook global texture information. Therefore, compared with the algorithm for detection and NMS, this algorithm is more robust to occlusion.
Last but not least, this method is independent of the number of faces. In the multi-face alignment task, bottom-up approaches will have a large margin than the top-down method in speed.

In this paper, we present a bottom-up algorithm that iteratively parses out a single face using global semantic segmentation information.
While our face task does not have clear connection like the limbs~\cite{cao2017realtime}, pixel embedding~\cite{de2017semantic} learns implicit features to obtain corresponding spatial feature relationships.
which compared with the top-down method which the detected faces are cyclically sent to the single-face landmarks network. 

In conclusion, our main contributions are threefold.
\begin{enumerate}
    \item We explored a bottom-up multiple face alignment structure, whose run-time is not correlated with the number of the face in an image.
    \item We proposed the Fox Block that can blend the global features and texture information of the face.
    \item We proposed a new loss function, Cosine Discriminative Loss, that introduces cosine function into the Discriminative Loss, which can classify facial features on high-dimensional space with better performance.
\end{enumerate}

The paper is organized as follows: Section 2 describes the proposed methods, Section 3 shows experimental results. Section 4 concludes this work.
\section{Proposed Method}
Figure.~\ref{fig:model_structure}A illustrates our bottom-up method.
% 1. 首先将整个算法的输入输出说清楚
The method takes an RGB image of size $w\times h$ and generates the landmarks and corresponding faces.
% 2. 将网络部分的输入输出说清楚
The FoxNet simultaneously predicts the landmark candidates $C$, at the segmentation branch, and their high-dimensional features $F$ which encode spatial information, at the feature branch.
% 3. Post-process的操作（infer和training不同） 
As shown in Figure.~\ref{fig:inference_structure}, features, which combine the non-maximum suppression result of segmentation branch, utilize cluster algorithm to produce multiple face landmarks.

\subsection{Architecture}
% 细讲FoxNet：1. Foxnet的组成（block层面） 2. 再讲我们的FoxBlock里面为什么要这么设计 
% 3. 最后说我们的训练的细节（feature branch训练时的是用的人脸instance里所有的像素，但infer时的点
% FoxNet is composed by the head and several stages of Fox Blocks. The input is first analyzed by a convolutional neural network, called head, e.g., ResNet~\cite{he2016deep} producing a set of the coarse feature map.

In our proposed networks, FoxNet, as illustrated in Figure~\ref{fig:model_structure} B., the first stage would produce a set of abstract feature $S^{1}=h(I)$, where $h$ are the head of FoxNet (e.g., ResNet~\cite{he2016deep}).
Moreover, in each subsequent stage, the block inherits multiscale information in the previous stage to produce more robust features $S^t = Fox(S^{t-1})$.
At the end, two  point-wise convolution of different the number of channel  generate segmentation result $C=d_1(S^t)$ and feature result $F=d_n(S^t)$, where $d$ is the depth-wise convolution and $n$ is the numebr of channel of them.
In order to fully utilize the facial multi-scale information, we view Hourglass~\cite{newell2016stacked} as our backbone. 
% 这里可以展开一下，具体说一下人脸有哪些局部特征和全局特征。
% 这里的fox block和fox net可以展开讲一下
Therefore, we designed a Fox Block that can blend multi-resolution identified features on the same scale.

\begin{figure}
    \centering
    \includegraphics[scale=0.24] {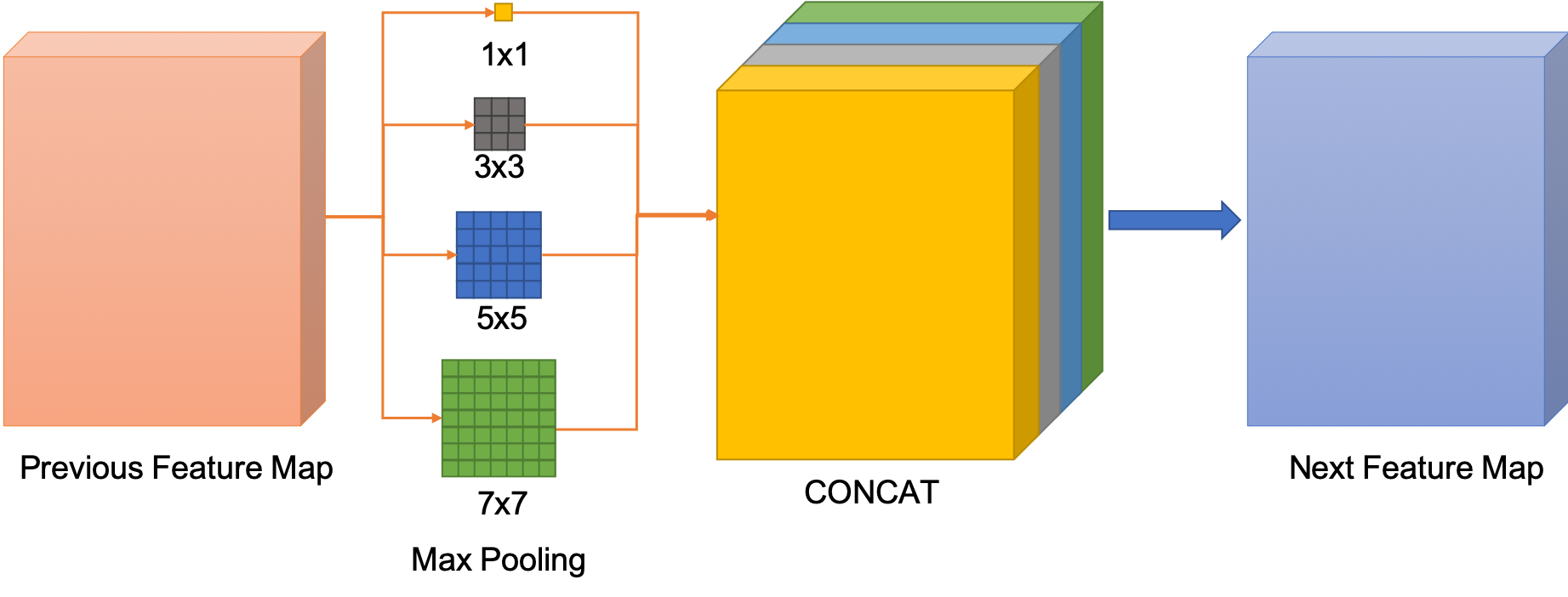}
    \caption{Four average poolings of different kernel harvest multiscale information and a point-wise convolution after concatenation correct to the previous channel size.\label{fig:foxblock}}
\end{figure}

% It takes four average pooling of different kernel size, $1, 3, 5, 7$, wand utilize $1\times 1$ convolution to fuse their concatenation result.
Our Fox Block, as shown in Figure.~\ref{fig:foxblock}, has four different kernel size, $1, 3, 5, 7$, of average pooling, which stride is $1$ to protect original resolution.
During inference, feature branch classifies landmark candidates come from segmentation branch.
However, during training time, as shown in Equation~\ref{equation:loss}, we make all facial pixel participated in the calculation to study more identified features.
\begin{equation}\label{equation:loss}
    l = L(P(I), T(P(I)))
\end{equation}
where $I$ is the input image, $P$ is pixel belonging to the face, $T$ is corresponding classification labels, and $L$ is our cosine loss.
% including global information, e.g., skin color and eye shape, and textual, e.g., expression and accessories.
% + Hourglass介绍
% The main networks structure is presented in Figure~\ref{fig:model_structure} B. We proposed to Hourglass~\cite{newell2016stacked} as the backbone. In Hourglass structure, features are processed across all scales and consolidated to best capture the various spatial relationships associated
% The networks use residual block~\cite{he2016deep}.

% \subsection{Fox Block}
% Fox Block applied to discover multi-scale detail features at each different resolution, followed by max average pooling with $kernel\;size=1,3,5,7, stride = 1$ and concatenation layers to form the final feature representation, which carries both local and global context information.

% \subsection{FoxNet}
% With Fox Block, we propose our FoxNet as illustrated in Figure~\ref{fig:model_structure} B. We proposed to Hourglass~\cite{newell2016stacked} as the backbone of our proposed networks. The reason for named the networks as FoxNet is that the shape of the structure of the network is like a fox.%Remain to be fixed
% In Hourglass structure, features are processed across all scales and consolidated to best capture the various spatial relationships associated
% The networks use residual block~\cite{he2016deep}.
% In order to extract the detail feature in the image, we would like to propose a structure that utilizes the Fox Block to extract the detail information.

To localize the landmark, the global information of the images is required. So we proposed to use Fox Block to have a larger receptive field in our proposed model.

% \subsection{ Fox Block}
% Fox Block applied to discover multi-scale detail features at each different resolution, followed by max average pooling with $kernel\;size=1,3,5,7, stride = 1$ and concatenation layers to form the final feature representation, which carries both local and global context information.
\subsection{Cosine Discriminative Loss}
% 开头说一下我们的任务和loss的联系
\textit{Pixel embedding}~\cite{de2017semantic} is a differentiable transform which maps each image pixels to high-dimensional vector for better classification.
The objective of our loss function is to increase the inter-class distance and minimize the intra-class distance.
% 然后是继承的哪个loss和它的特点
Discriminative loss~\cite{de2017semantic} has made great success in semantic segmentation field which enforces the network to map each pixel in the image to an n-dimensional vector in feature space. However, we viewed that introducing the cosine loss takes the normalized features as input to learn highly discriminative features by maximizing the inter-class cosine margin could utilize the cosine-related discriminative information well.
\cite{de2017semantic} uses variance term to force embeddings to close the cluster center, distance term to push away the cluster centers from each other and regularization to pull all embeddings to the origin.
% 接着描述它的不足和我对它的改进
% 1. 因为我们不需要向量的长度，只需要向量的方向（这句话该怎么描述：向量的长度代表着该像素对关键点的响应值（因为在segementation的分支里已经预测了关键点了，这里就不需要这个信息了），方向才是我们需要的特征信息？？）
% 2. 第三个term让len(vector）强制规约到原点。因为我们需要的是方向，不是长度。所以我们让长度规约到一个常数上（refer ring loss）。
% 3. 参考ring loss的可视化图，最终的点都是在球表面的上下波动，所以直接使用l2 loss不能代表真正的特征距离，所以我们使用计算两个向量的cosine值
We inherit three-terms, but replace the Euclidean distance with cosine distance and change the pull to the push. % 这里想表达的意思是我们要把向量推远
In our task, we only need the orientation of embedding to obtain the discriminative features.
As shown in figure.~\ref{fig:model_structure}A, our segmentation predicts the landmark candidates who have more precise semantic information than the length of embedding who represent the response of landmark on a feature branch.
If we use regularization term, in discriminative loss, forcing embeddings of different length into the origin, the surface area of the characteristic hypersphere will too small to classify.
Inspired by~\cite{zheng2018ring}, we put embeddings to a hypersphere with extensive surface area which can learn better distribution and normalization to cluster.

% The length of embedding represents the response of landmark and direction.
In cosine discriminative loss, regularization term force embeddings of the different norm into the origin, which make the surface area of the characteristic hypersphere shrank.  
The details of our proposed Fox Loss is illustrate in  Equation~\ref{equation:L_final} to maximizing inter-class variance and minimizing intra-class variance. It has integrated Equation~\ref{equation:L_var} to~\ref{equation:L_reg}.
The variance term($L_{var}$) is an intra-cluster pull-force that draws embeddings towards the mean embedding which has presented in Equation~\ref{equation:L_var}. The distance term is an inter-cluster push-force that pushes clusters away from each other, increasing the distance between the cluster centers which has presented in Equation~\ref{equation:L_dist}. The regularization term is a small pull-force that draws all clusters towards the origin, to keep the activations bounded which has presented in Equation~\ref{equation:L_reg}. In the equations, the definitions are as follows:
$C$ is the number of clusters in the ground truth, $N_c$ is the number of elements in cluster $c$, $x_i$ is an embedding, $\mu_c$ is the mean embedding of cluster $c$ (the cluster center), 
% $||\cdot||$ is the L1 or L2 distance, j∫
$cosine(a,b)$ is the cosine loss between $a$ and $b$, which could also be noted as $\frac{a \cdot b}{||a||\cdot ||b||}$.
$[x]_+=max(0,x)$ denotes the hinge. $\delta_v$ and $\delta_d$ are respectively the margins for the variance and distance loss.

our cosine discriminative loss is defined as follows:
\begin{equation}\label{equation:L_var}
    L_{var}=\frac{1}{C}\sum^C_{c=1}\frac{1}{N_c}\sum^{N_c}_{i=1}\left[cosine(\mu_c,x_i)-\delta_v\right]^2_+
\end{equation}
\begin{equation}\label{equation:L_dist}
    L_{dist}=\frac{1}{C(C-1)}\sum^C_{c_A=1}\sum^C_{c_B=1}\frac{1}{N_c}\sum^{N_c}_{i=1}\left[2\delta_d-cosine(\mu_{cA},\mu_{cB})\right]^2_+
\end{equation}
\begin{equation}\label{equation:L_reg}
    L_{reg}=\frac{1}{C}\sum^C_{c=1}(||\mu_c||_2 - R)^2
\end{equation}
\begin{equation}\label{equation:L_final}
    L_{fox}=\alpha \cdot L_{var}+\beta \cdot L_{dist}+\gamma \cdot L_{reg}
\end{equation}

\subsection{Semi-supervised Face Separation with Mean Shift}
Different from the traditional structure that iteratively passes the facial information into the prediction networks, all facial information has been presented on our segmentation branch. 
The corresponding facial landmarks share some particular feature. Landmarks that belong to the same face can be seen as a cluster in Euclidean space. For example, the Euclidean distance of each landmark is closer to other faces.
Mean shift~\cite{comaniciu2002mean} is a procedure for locating the modes of a density function given discrete data sampled from that function which involves shifting this kernel iteratively to a higher density region until convergence. It always points toward the direction of the maximum increase in the density. 
The complexity will tend towards $O(T*n*log(n))$ in lower dimensions, with $n$ the number of samples and $T$ the number of points. It is suitable for a mean shift to process clustering on facial landmarks. It is a semi-supervised clustering algorithm that allows the input without given the number of clusters.
We perform a mean shift algorithm to separate the corresponding face information.
\begin{figure}
    \centering
    \includegraphics[scale=0.3]{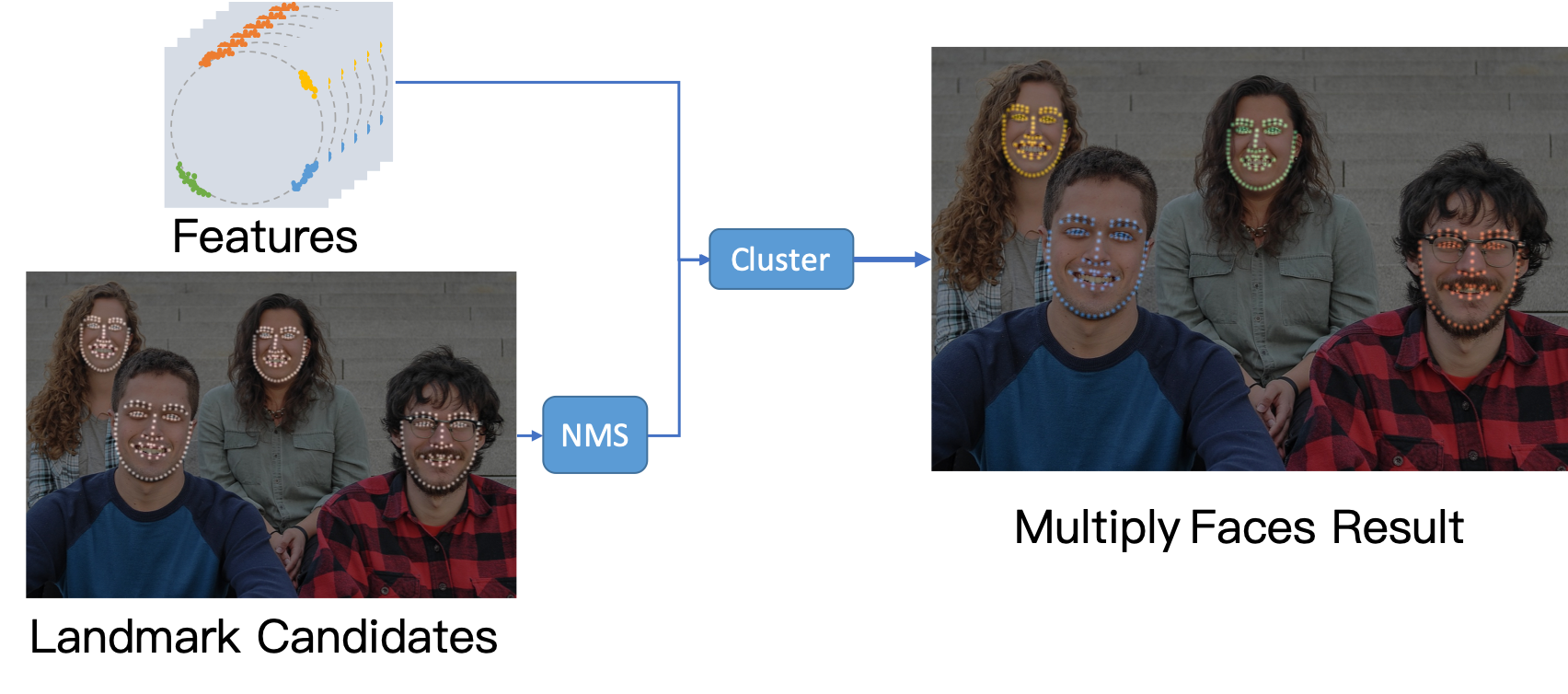}
    \caption{In inference, we obtain the landmark result after NMS which utilize its high-dimensional feature to cluster and parse to the multiple faces.
    The color of Landmarks Candidates is the same but Muliply Faces Results are different\label{fig:inference_structure}}
\end{figure}
As presented on Figure~\ref{fig:inference_structure}, in our inference. We perform non-maximum suppression(NMS) operation on the segmentation branch from the training process and utilize the results to perform the mean shift to separate the different faces.
We utilize the segmentation branch from the training process and perform NMS operation.

\section{Experiments}
We evaluate our method on two datasets: Single Face Dataset WFLW and our Multi-face AISA Dataset for precision and speed.
\begin{table*}[!ht]
    \centering
    \begin{tabular}{|c|c|c|c|c|c|c|c|c|}
        \hline
        Method&FullSet&Pose&Expression&Illumination&Makeup&Occlusion&Blur\\\hline
        ESR~\cite{cao2014face}&11.13&25.88&11.47&10.49&11.05&13.75&12.20\\\hline
        CFSS~\cite{zhu2015face}&9.07&21.36&10.09&8.30&8.74&11.76&9.96\\\hline
        LAB~\cite{wu2018look}&\textbf{5.27}&\textbf{10.24}&\textbf{5.51}&\textbf{5.23}&\textbf{5.15}&6.79&6.32\\\hline
        \textbf{OURS}&5.80&10.50&8.94&5.71&6.30&\textbf{6.53}&\textbf{6.30}\\\hline
    \end{tabular}
    \caption{The Experiment Results Meassured on NME(\%) on WFLW Dataset}\label{exp:WFLW}
\end{table*}
% \textbf{Dataset.} We conduct evaluation on four challenging datasets including WFLW~\cite{wu2018look} and our AISA Dataset.

\textbf{WFLW dataset}: WFLW contains 10000 faces(7500 for training and 2500 for testing) with $98$ munually annotated landmarks. 

\textbf{AISA Dataset}: In order to facilitate the bottom-up multi-face alignment algorithm, we introduce a new dataset base
on 300W~\cite{sagonas2016300}, which contains $3000$($2500$ for training and $500$ for testing).
The difficulty is reflected in face scale, occlusion and the number of faces.

\textbf{Evaluation metric.} We use standard normalized landmarks mean error(NME) to evaluate face landmarks
moreover, the F1 score to evaluate face detection.

\subsection{Evaluating Single Face Alignment}
We compare our method against the state-of-the art methods, ESR~\cite{cao2014face}, CFSS~\cite{zhu2015face} and LAB~\cite{wu2018look}, on WFLW. 
The result is shown in Table~\ref{exp:WFLW} which  comes from segmentation branch using NMS.

Our method achieves $5.80\%$ on the test set and higher than LAB, while better on Occlusion and Blur subset.
This margin shows that our method has a larger receptive field to obtain more global features.

\subsection{Comparing Bottom-up and Top-down Method}
Top-down multi-face alignment method contains detection and single face alignment, so we compare our approach
with these two steps, respectively. As shown in Table ~\ref{exp:time},

\begin{table}[!ht]
    \centering
    \begin{tabular}{|c|p{60pt}|c|c|}
        \hline
        Detection Method&Single Face Alignment Method&F1 Score&NME\\\hline
        MTCNN&LAB&0.56&\textbf{6.10}\\\hline
        SSH&LAB&\textbf{0.89}&\textbf{6.10}\\\hline
        \textbf{OURS}&\textbf{OURS}&0.80&6.80\\\hline
    \end{tabular}
    \caption{Our model compare them with detection and single alignment respectively on our dataset\label{exp:time}}
\end{table}

Our detection result $0.80\%$ is better than MTCNN~\cite{zhang2016joint}, 
and the alignment result $6.80\%$ is slightly worse than LAB~\cite{wu2018look}

\subsection{Runtime Analysis}
\begin{figure}
    \centering
    \includegraphics[scale=0.34]{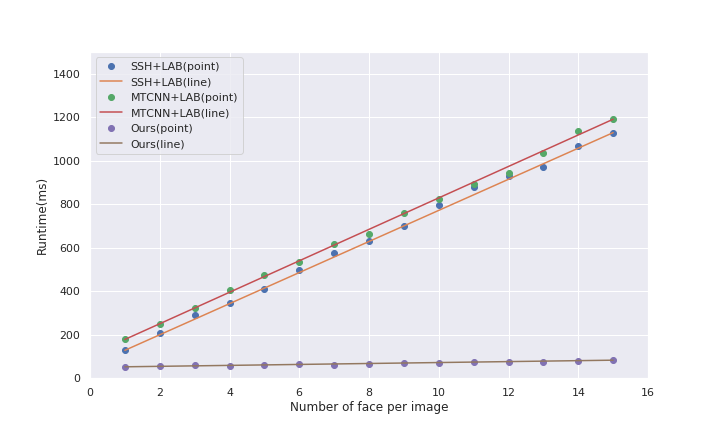}
    \caption{Runtime Analysis\label{fig:speed}}
\end{figure}

To analyze the runtime performance of our method, we uniformly resize to $640\times 640$ during test time to fit GPU memory.
The runtime analysis is performed on a single NVIDIA GeForce GTX-1080ti GPU. 
We perform face detection SSH~\cite{najibi2017ssh} and two single face DAN~\cite{iyyer2015deep} and LAB as a top-down comparison, where the runtime is roughly proportional to the number of people in an image. The results are illustrated in Fig.~\ref{fig:speed}. In our approach, we only took $51.50\; ms$ to process the single face landmark detection task while the baseline experiments that perform on SSH+LAB and MTCNN+LAB would take $127.34\; ms$ and $177.259\;ms$. Compared to the other two methods, the slope of our proposed method is minimal(to be more precise, our slope is $2.06$ and the slope of the other two is $71.65$ (SSH+LAB) and $71.83$ (MTCNN+LAB).). It is obvious our proposed method is not only the fastest in single face alignment task but is increases relatively slowly with the increasing number of people. The runtime consists of two major parts:
\begin{enumerate}
    \item In our structure, CNN only processed once which is constant with varying number of people; 
    \item Multi-face parsing time whose runtime complexity is $O(n*log(n))$, where $n$ is represents the number of faces.
    However, the parsing time does not significantly influence the overall runtime because it is one order of magnitude less than the CNN processing time, e.g., for $9$ people, the parsing time takes $5.54\; ms$ while CNN takes $52\;ms$.
\end{enumerate}

% We directly calculate the heatmap result of the semantic segmentation branch using the NMS of the point to calculate the corresponding coordinates.

\subsection{Training Details}
All models are implemented using PyTorch~\cite{paszke2017automatic} and trained on a GPU server with $8$ NVIDIA GTX 1080Ti GPU. The training details here are all similar to that in~\cite{wu2018look}. To facilitate future research and clarify the details. Some important training details are as follows:

\begin{enumerate}
    \item Learning Rate: $0.5$
    \item Epoch of warm-up : $3$
    \item Epoch : $140$
    \item Optimizer : SGD
    \item Random Crop : $640 \times 640$
    \item Batch Size = $1$
\end{enumerate}
We set $\alpha = \beta =1$, $\gamma = 0.001$ and $\delta_v = \delta_d = 1$~\cite{de2017semantic}. 
% Table~\ref{exp:WFLW} depicts the performance of the proposed models and their corresponding baselines(the lower the better). This demonstrates that our proposed method, has aced on Occulusion and Blur on every baseline results and we maintain the close results to the state-of-art result in~\cite{wu2018look}.

\section{Conclusion}
We have developed an extremely fast structure that develops the multi-face alignment task. It is the first bottom-up structure on this task.

In our approach, we first proposed the use of the FoxNet structure to solve the problem of receptive field defects. Moreover, we use Fox Block to provide additional contextual information that may be needed for facial landmark detection. In our approach, we have achieved a high-speed bottom-up solution and maintain most of the accuracy. The approach is an algorithm that is independent of the number of people to be detected which could be applied on large-scale real-time facial alignment task.

\section{Acknowledgments}
This reseach was supported in part by the National Key Technology R\&D Program of China $(2017$YFD$0301506)$ and the Hunan
Province innovative experiment and reseach study program for college student $($SCX$1822)$.
\bibliographystyle{IEEEbib}
\bibliography{refer}
\end{document}